\pdfoutput=1
\PassOptionsToPackage{table}{xcolor}
\documentclass[11pt]{article}

\usepackage[preprint]{acl}

\usepackage{times}
\usepackage{latexsym}

\usepackage{bm}
\usepackage[table]{xcolor}
\usepackage{booktabs} 
\usepackage{multirow}
\usepackage[ruled,linesnumbered,vlined]{algorithm2e}
\usepackage{amsmath}
\usepackage{amssymb}
\usepackage{mathtools}
\usepackage{amsthm}
\usepackage[subrefformat=parens,labelformat=parens]{subfig}
\usepackage{lipsum}
\usepackage{wrapfig}
\usepackage{makecell}

\usepackage[T1]{fontenc}

\usepackage[utf8]{inputenc}

\usepackage{microtype}

\usepackage{graphicx}

\title{AROMA: Autonomous Rank-one Matrix Adaptation}

\author{
 \textbf{Hao Nan Sheng\textsuperscript{1}},
 \textbf{Zhi-yong Wang\textsuperscript{1}},
 \textbf{Mingrui Yang\textsuperscript{2,3}},
 \textbf{Hing~Cheung~So\textsuperscript{1}},
\\
\\
 \textsuperscript{1}City University of Hong Kong
 \textsuperscript{2}The University of Hong Kong\\
 \textsuperscript{3}AI Chip Center for Embedded Smart System
 \\
 {
   \href{mailto:hnsheng2-c@my.cityu.edu.hk}{hnsheng2-c@my.cityu.edu.hk}
 }
}

\begin{document}
\maketitle
\begin{abstract}

As large language models continue to grow in size, parameter-efficient fine-tuning (PEFT) has become increasingly crucial. While low-rank adaptation (LoRA) offers a solution through low-rank updates, its static rank allocation may yield suboptimal results. Adaptive low-rank adaptation (AdaLoRA) improves this with dynamic allocation but remains sensitive to initial and target rank configurations. We introduce AROMA, a framework that automatically constructs layer-specific updates by iteratively building up rank-one components with very few trainable parameters that gradually diminish to zero. Unlike existing methods that employ rank reduction mechanisms, AROMA introduces a dual-loop architecture for rank growth. The inner loop extracts information from each rank-one subspace, while the outer loop determines the number of rank-one subspaces, i.e., the optimal rank. We reset optimizer states to maintain subspace independence. AROMA significantly reduces parameters compared to LoRA and AdaLoRA while achieving superior performance on natural language understanding and commonsense reasoning tasks, offering new insights into adaptive PEFT. The code is available at \href{https://github.com/ShuDun23/AROMA}{AROMA}.
\end{abstract}

\section{Introduction}



The emergence of large language models (LLMs) \citep{devlin2019bert, openai2023gpt, llama3, liu2024deepseek} has revolutionized the field of natural language processing (NLP), yet their full potential is often limited by the substantial computational demands of fine-tuning. Traditional full-parameter tuning, while effective, becomes prohibitively expensive as model sizes escalate into hundreds of billions of parameters \citep{lester2021power,meng2024pissa}. For instance, LLaMA3 series boasts models with up to 400B parameters \citep{meta2024introducing}, and DeepSeek-V3 encompasses 671B total parameters due to its mixture-of-experts architecture \citep{liu2024deepseek}. This challenge has driven the development of parameter-efficient fine-tuning (PEFT) methods, such as prompt-tuning \citep{lester2021power}, prefix-tuning \citep{liliang2021prefix}, and adapter tuning \citep{pfeiffer2020adapterfusion, houlsby2019parameter}. Besides these, low-rank adaptation (LoRA) \citep{hu2022lora} stands out as a particularly promising approach for its simplicity and strong theoretical foundation.

LoRA learns incremental low-rank update $\varDelta \bm{W}$ to pretrained model $\bm{W}_0$, without altering the model architecture or introducing additional inference latency \citep{hu2022lora}. While attaining impressive parameter efficiency (typically less than 1\% of full fine-runing), conventional LoRA implementations impose uniform rank allocation across all layers. This might be suboptimal, as different components of the network exhibit varying sensitivities to parameter perturbations \citep{zhang2023adalora}. Moreover, determining the optimal ranks remains an empirical process that often necessitates extensive trial-and-error experimentation. 





As a modified version, adaptive low-rank adaptation (AdaLoRA) \citep{zhang2023adalora} adopts dynamic rank allocation through singular value decomposition (SVD)-based importance scoring. While it improves the flexibility upon static configurations like LoRA, it still faces several limitations: 1) the need to prespecify both the initial and target rank budgets; 2) substantial computational overhead caused by relaxed SVD; and 3) rank redundancy stemming from a low effective rank proportion. Consequently, the fundamental tension between adaptive rank adjustment and computational efficiency remains an open question.



\begin{figure*}[t]
  \centering
  \subfloat[\#Parameter]
  {
      \label{subfig param vs step}\includegraphics[width=0.19\linewidth]{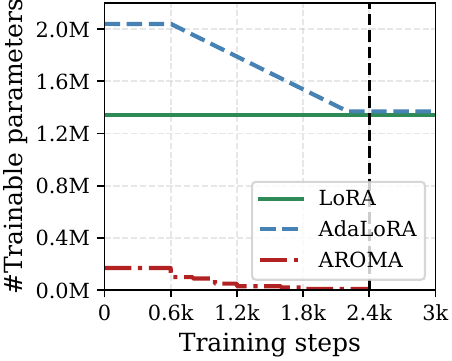}
  }
  \centering
  \subfloat[Total rank]
  {
      \label{subfig tot tank vs step}\includegraphics[width=0.19\linewidth]{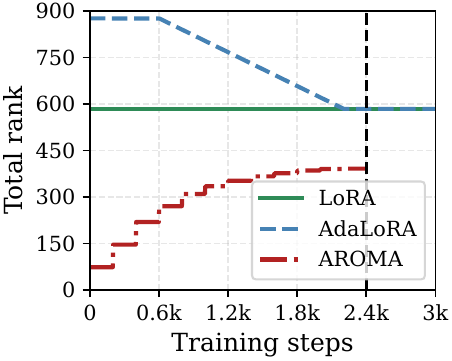}
  }
  \centering
  \subfloat[Specific rank]
  {
      \label{subfig rank vs step 1}\includegraphics[width=0.19\linewidth]{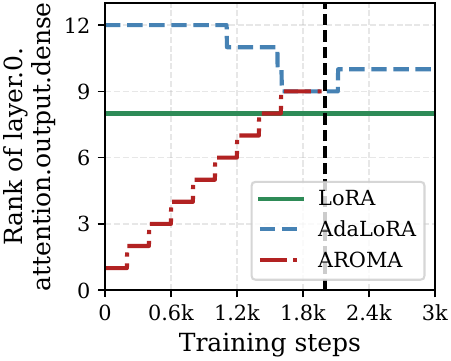}
  }
  \centering
  \subfloat[Specific rank]
  {
      \label{subfig rank vs step 2}\includegraphics[width=0.19\linewidth]{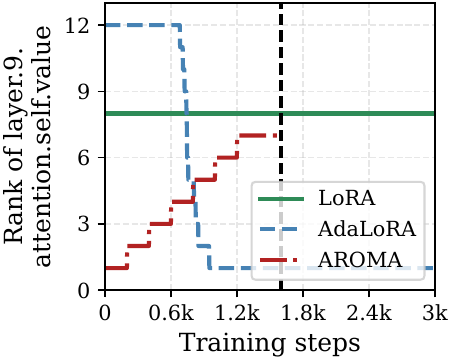}
  }
  \centering
  \subfloat[Accuracy]
  {
      \label{subfig acc vs step}\includegraphics[width=0.19\linewidth]{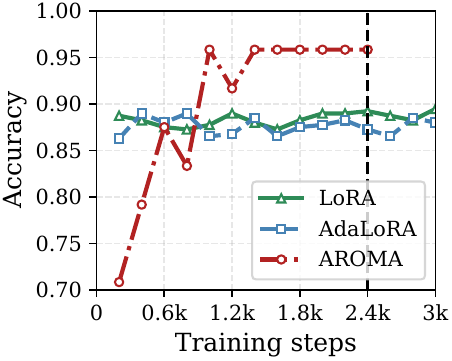}
  }  
  \caption{Results for LoRA\textsubscript{$r$=8}, AdaLoRA\textsubscript{$r$=8}, and AROMA (ours) include the number of trainable parameters, total rank, rank of a specific layer and evaluation accuracy versus training step for RoBERTa-base on MRPC task. For AROMA, training of "layer.0.attention.output.dense" and "layer.9.attention.self.value" automatically terminates at 2000 and 1600 steps, respectively, while the overall training automatically stops at 2400 steps.}
  \vspace{-3mm}
  \label{fig: training steps}
\end{figure*}

In this work, we present \textbf{A}utonomous \textbf{R}ank-\textbf{O}ne \textbf{M}atrix \textbf{A}daptation (AROMA), a novel rank-growing low-rank adaptation method that reconsiders the dynamics of rank allocation. Experimental results demonstrate that AROMA significantly outperforms both LoRA and AdaLoRA when applied to the RoBERTa-base \citep{liu2019roberta} on the GLUE benchmark \citep{wang2018glue} and the LLaMA3-8B \citep{llama3} on the commonsense170K dataset \citep{hu2023llm}. Notably, AROMA achieves this enhanced performance only using <10\% of the parameters required by LoRA\textsubscript{$r$=8} and AdaLoRA\textsubscript{$r$=8} without prespecified rank.
Main contributions are summarized as follows: 
\begin{itemize}
    \item \textbf{Adaptive Rank Growth} We propose a structure that progressively establishes layer-specific ranks with minimal and decreasing trainable parameters. Unlike AdaLoRA's pruning-based strategy, AROMA initiates with zero rank and incrementally incorporates rank-one components until convergence criteria are met. This bottom-up structure ensures high parameter efficiency without loss of informative subspaces. \vspace{-1mm}
    \item \textbf{Automatic Rank Convergence} AROMA features a dual-loop architecture for automatic rank control. Each module operates with an inner loop that extracts information from individual rank-one subspace, and an outer loop determines the number of these subspaces, i.e., the optimal rank. We design a convergence criterion for both loops, enabling each module to autonomously determine the appropriate rank without the need to predefine it. \vspace{-1mm}
    \item \textbf{Independent Subspace} We introduce a training strategy termed \emph{Check \& Merge \& Reinit \& Reset}, which includes convergence checking, merging converged rank-one updates, periodic optimizer resets alongside learning rate warmup. After each inner loop, the optimizer states are reset while preserving the knowledge accumulated in the weights. This facilitates subspace switching, leading to high effective rank proportion and a continuous flow of new domain knowledge.
\end{itemize}

\section{Background and Motivation}

\begin{figure*}[t]
    \centering
    \includegraphics[width=0.8\linewidth]{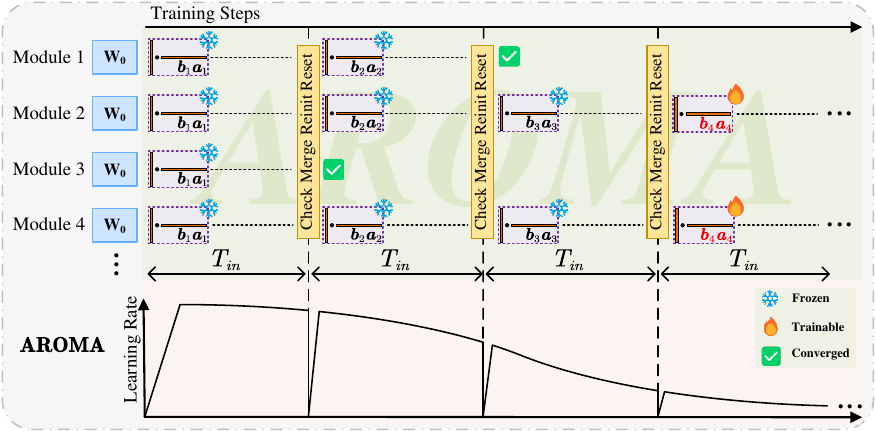}
    \caption{Workflow of \textit{\textbf{AROMA}}.
    For each module, AROMA trains rank-one matrices sequentially with a dual-loop architecture. In the inner loop, a rank-one LoRA, $\bm{b}\bm{a}$, is updated, whose convergence is assessed by the inner stopping criterion. Prior to heading to next outer loop step, we check outer convergence by outer stopping criterion. If not converged, the computed rank-one components are merged and frozen, and new $\bm{b}$ and $\bm{a}$ are initialized for training with reset learning rate and optimizer states. For simplicity, we illustrate the length of inner loop to $T_{\mathrm{in}}$, though in practice, it is determined by both $T_{\mathrm{in}}$ and the inner convergence criterion.}
    \vspace{-3mm}
    \label{fig:AROMA}
\end{figure*}

LoRA \citep{hu2022lora} fine-tunes the pretrained model $\bm{W}_0 \in \mathbb{R}^{m\times n}$ by incorporating a low-rank decomposition, namely:
\begin{equation}
    \bm{W} = \bm{W}_0 + \frac{\alpha}{r}\varDelta\bm{W}, \ \varDelta\bm{W}=\bm{B}\bm{A}
\end{equation}
where $\bm{B} \in \mathbb{R}^{m \times r}$, $\bm{A}\in \mathbb{R}^{r \times n}$ with $r \ll \min\{m,n\}$, and scaling factor $\alpha$ secures consistent output magnitude across different rank values. 
However, this approach requires careful selection of $r$ and imposes uniform rank across all layers, potentially not optimal.

AdaLoRA \citep{zhang2023adalora} addresses these static allocation limitations by parameterizing the incremental matrix as $\bm{P\varLambda Q}$, mimicking SVD while enforcing orthogonality:
\begin{equation}
\begin{aligned}
    \varDelta\bm{W} &= \ \bm{P}\bm{\varLambda}\bm{Q}, \\ 
    \mathrm{s.t.}\ \bm{P}^T\bm{P}&=\bm{Q}\bm{Q}^T=\bm{I}_r
\end{aligned}
\end{equation}
where $\bm{P}\in \mathbb{R}^{m\times r}$ and $\bm{Q}\in \mathbb{R}^{r\times n}$ represent left and right singular vectors while $\bm{\varLambda}\in \mathbb{R}^{r\times r}$ stores singular values. AdaLoRA begins with a high initial total rank budget and gradually reduces it at certain intervals. Specifically, singular values across all layers are sorted in descending order based on the importance score, with only the top $b^{(t)}$ retained, ultimately converging to a target rank budget. Since these singular values belong to different module weights, this mechanism enables adaptive rank allocation across modules. Nevertheless, AdaLoRA exhibits several limitations:
\begin{itemize}
    \item Like LoRA, AdaLoRA's performance remains sensitive to the initial and target total rank configurations. Optimal rank selection is task-dependent and architecture-specific, complicating deployment in empirical scenarios.
    \item Computing the relaxed SVD in AdaLoRA introduces substantial complexity that scales linearly with layer dimensions, creating computational bottlenecks for very large models.
    \item The higher initial ranks demand substantial memory allocation during early training phases, imposing practical limitations in resource-constrained environments.
\end{itemize}



Against these backdrops, we devise an automatic and adaptive rank-growing scheme inspired by rank-one matching pursuit \cite{wang2014rank, wang2015orthogonal}. This approach leverages the principle that any rank-$r$ matrix $\bm{L}$ can be decomposed into a sum of $r$ rank-one matrices:
\begin{equation}
    \bm{L}=\sum_{p=1}^{r}{\bm{b}_p\bm{a}_{p}}
\end{equation}
where $\bm{b}_p\in \mathbb{R}^{m\times 1}$ and $\bm{a}_p\in \mathbb{R}^{1\times n}$. Building on this idea, we develop our novel framework.




\section{Methodology}
This section outlines two crucial aspects of AROMA: 1) the adaptive rank-growing mechanism, featuring both inner and outer stopping criteria; and 2) the training strategy known as Check \& Merge \& Reinit \& Reset. Figure \ref{fig:AROMA} depicts the AROMA framework, and Algorithm \ref{Alg1} in Appendix \ref{Apdx: alg1} provides the detailed steps.

\subsection{Adaptive Rank Growth}


Unlike AdaLoRA that truncates singular values with low important scores, we propose a rank-growing scheme which introduces a dual-loop training structure: the inner loop computes individual rank-one matrix, while the outer loop determines the quantity of these matrices. For the $p$th outer loop step, $\varDelta\bm{W}$ is parameterized as:
\begin{equation}\label{deltaW}
\begin{split}
    \varDelta\bm{W}&=\bm{b}_1\bm{a}_1 + \bm{b}_2\bm{a}_2 + \cdots + \bm{b}_{p-1}\bm{a}_{p-1} + \textcolor{red}{\bm{b}_{p}\bm{a}_{p}}\\
    &=\left[ \begin{matrix}
	\bm{B}_{p-1}&		\textcolor{red}{\bm{b}_p}\\
    \end{matrix} \right] \left[ \begin{array}{c}
	\bm{A}_{p-1}\\
	\textcolor{red}{\bm{a}_p}\\
    \end{array} \right] 
\end{split}
\end{equation}
where $\bm{B}\in\mathbb{R}^{m\times p}$ and $\bm{A}\in\mathbb{R}^{p\times n}$. 

\textbf{AROMA learns a series of rank-one LoRAs.} At the beginning of the $p$th outer iteration, a new rank-one LoRA $\bm{b}_p\bm{a}_p$ is activated for training, while previously calculated $\bm{b}_1\bm{a}_1, \bm{b}_2\bm{a}_2, \cdots, \bm{b}_{p-1}\bm{a}_{p-1}$ are frozen and merged as a single matrix $\bm{B}_{p-1}\bm{A}_{p-1}$. 

Next, $\bm{b}_p^{(0)}$ and $\bm{a}_p^{(0)}$ enter the inner loop. Here we denote the update in the $t$th inner loop step as $\bm{b}_p^{(t)}$ and $\bm{a}_p^{(t)}$. They update until $t$ reaches the maximum inner steps $T_{\mathrm{in}}$ or the inner stopping criterion is met:
\begin{equation}\label{check_in}
    \frac{\left\| \boldsymbol{b}_p^{(t)}\boldsymbol{a}_p^{(t)} \right\| _F - \left\| \boldsymbol{b}_p^{(t-\varDelta T_{\mathrm{in}})}\boldsymbol{a}_p^{(t-\varDelta T_{\mathrm{in}})} \right\| _F}{\left\| \boldsymbol{b}_p^{(t-\varDelta T_{\mathrm{in}})}\boldsymbol{a}_p^{(t-\varDelta T_{\mathrm{in}})} \right\| _F}<\varepsilon_{\mathrm{in}}
\end{equation}
where $\varepsilon_\mathrm{in}$ denotes the inner convergence tolerance, and $\varDelta T_{\mathrm{in}}$ is the inner checking interval. We evaluate \eqref{check_in} every $\varDelta T_{\mathrm{in}}$ steps, and if it is satisfied, the inner loop terminates, and the training of $\bm{b}_p\bm{a}_p$, viz., current rank-one LoRA, is completed.

\textbf{When to stop?} Once the inner loop ends, we check for outer loop convergence before proceeding to the next outer loop step. Here we use a relative weight change criterion between the $(p-1)$th and the $p$th outer steps defined as:
\begin{equation}\label{check_out}
\begin{aligned}
    &\frac{\left\| \left( \boldsymbol{W}_0 + \alpha \boldsymbol{B}_p\boldsymbol{A}_p \right) - \left( \boldsymbol{W}_0 + \alpha\boldsymbol{B}_{p-1}\boldsymbol{A}_{p-1} \right) \right\| _F}{\left\| \boldsymbol{W}_0+\alpha \boldsymbol{B}_{p-1}\boldsymbol{A}_{p-1} \right\| _F} \\
    =&\frac{\left\| \alpha \boldsymbol{b}_p\boldsymbol{a}_p \right\| _F}{\left\| \boldsymbol{W}_0+\alpha \boldsymbol{B}_{p-1}\boldsymbol{A}_{p-1} \right\| _F}
    <\varepsilon_\mathrm{out} 
\end{aligned}
\end{equation}
where $\varepsilon_\mathrm{out}$ denotes the outer convergence tolerance. If \eqref{check_out} is satisfied, the outer loop will terminate, viz., training of $\varDelta\bm{W}$ is completed. 

Since we only leverage rank-one updates, each update can be regarded as a basis spanning a rank-one matrix subspace, which encompasses different domain knowledge. In AROMA, the inner loop exploits each subspace, yielding a rank-one basis $\bm{b}_p^{(t)}\bm{a}_p^{(t)}$, while the outer loop continuously pursues new subspaces and determines the appropriate number of subspaces. This rank-growing strategy allows for continuously extraction new information while keeping only one rank-one matrix trainable at a time, securing high parameter efficiency.

Furthermore, we implement AROMA across all modules, and train them in parallel (see Figure \ref{fig:AROMA}). For the inner loop, each module has its own inner convergence label and advances to the next outer step when all modules have either converged or reach $T_{\mathrm{in}}$. In particular, the module that converges will continue training while waiting for the others to catch up prior to proceeding together to the next outer step. Apart from facilitating rank allocation, this approach helps prevent premature termination, ensuring a more comprehensive subspace exploration. 

On the other hand, each module also possesses an outer convergence label, and once a module is determined as converged according to \eqref{check_out}, it is immediately frozen and the latest rank-one component will not be merged into it, while training continues for the remaining modules. The overall training process finishes when all modules converge or reach the maximum total training steps $T$. This design allows each module to determine the optimal rank independently and autonomously, enabling adaptive rank growth with a gradually reduced trainable parameters. We list the time complexity of LoRA, AdaLoRA and AROMA in Table \ref{table: complexity}, where $\tilde{r}$ denotes the current rank for AdaLoRA. Typically, we have $\mathcal{O}_{\text{AdaLoRA}} > \mathcal{O}_{\text{LoRA}} \geq \mathcal{O}_{\text{AROMA}}$. Detailed analyses and experimental verification are presented in Appendix \ref{Apdx: complexity} and Section \ref{efficiency}, respectively.
\begin{table}[ht]
    \centering
    \fontsize{8}{12}\selectfont 
    \setlength{\tabcolsep}{4pt} 
        \begin{tabular}{c | c c c}
        \toprule[1pt] \rowcolor{cyan!10}
        \bf Scheme & LoRA & AdaLoRA & \textit{\textbf{AROMA}} \\
        \midrule
        \bf Complexity & $\mathcal{O}((m+n)r)$ & {$\mathcal{O}((m+n)\tilde{r})$} & \cellcolor{gray!10} $\mathcal{O}((m+n)p)$ \\
        \bottomrule[1pt]
        \end{tabular}
    \caption{Per-step complexity comparison}
    \vspace{-4mm}
    \label{table: complexity}
\end{table}

\subsection{Check \& Merge \& Reinit \& Reset}
We further design a training strategy known as \emph{Check \& Merge \& Reinit \& Reset}. As its name implies, there are four components.

\noindent \textbf{Check} involves the inner and outer convergence criteria described in \eqref{check_in} and \eqref{check_out}. The inner checks occur every $\varDelta T_{\mathrm{in}}$ steps, while the outer checks take place when the inner loop finishes. 

\noindent \textbf{Merge \& Reinit} where \emph{Reinit} stands for reinitialize. As mentioned before, if \eqref{check_out} is met, we terminate the outer loop. Otherwise, the previously computed $\bm{b}_p\bm{a}_p$ is merged into $\bm{B}_{p-1}\bm{A}_{p-1}$, and the training progresses to the next outer step. At this point, a new rank-one LoRA $\bm{b}_{p+1}\bm{a}_{p+1}$ is introduced, with Kaiming initialization \citep{he2015delving} for $\bm{a}_{p+1}^{(0)}$ and zero for $\bm{b}_{p+1}^{(0)}$.

\noindent \textbf{Reset} represents optimizer state reset. 
With momentum parameters $\beta_1=0.9$ and $\beta_2=0.999$, Adam optimizer \citep{kingma2014adam, loshchilov2018decoupled} tends to follow established optimization paths, as update steps are strongly influenced by previous gradients.
This means that after Merge \& Reinit, the previous updates still influence current learning, causing the new LoRA update to continue exploring the learned subspaces. To circumvent this,
we randomly prune 99.9\% of the optimizer states following each Merge \& Reinit. Such an idea of subspace switching is adopted in LLM pretraining \citep{lialin2023relora,zhao2024galore} and subspace learning \citep{larsen2021many,gur2018gradient}. 

Additionally, a warmup phase is implemented at the start of training for each LoRA update to mitigate early overfitting. While the initial warmup phase is set to hundreds of steps, subsequent quick warmup phases are limited to tens of steps. The learning rate scheduler is illustrated in Figure \ref{fig:AROMA}.

\section{Experiments}
\begin{table*}[t]
    \centering
    \fontsize{9}{13}\selectfont 
    \setlength{\tabcolsep}{5pt} 
        \begin{tabular}{c|c|cccccccc|c}
        \toprule[1pt] \rowcolor{cyan!10}
        & & \textbf{\texttt{CoLA}} & \textbf{\texttt{MNLI}} & \textbf{\texttt{MRPC}} & \textbf{\texttt{QNLI}} & \textbf{\texttt{QQP}} & \textbf{\texttt{RTE}} & \textbf{\texttt{SST-2}} & \textbf{\texttt{STS-B}} & \\ \rowcolor{cyan!10}
        \multirow{-2}{*}{\bf Scheme} & \multirow{-2}{*}{\bf \#Param} & MC & Acc & Acc & Acc & Acc & Acc & Acc & PC & \multirow{-2}{*}{\textbf{Avg}} \\
        \midrule
        Full Fine-tuning & 125.0M & 60.26 & \bf87.68 & 88.33 & 92.58 & 90.75 & 78.63 & 94.63 & 90.31 & 85.40 \\ 
        \midrule
        BitFit\textsuperscript{$\sharp$} & 0.10M & 61.16 & 85.50 & 89.07 & 90.99 & 88.08 & 79.57 & 94.38 & 90.55 & 84.91 \\ 
        \midrule
        Adapter\textsuperscript{H}\textsuperscript{$\dagger$} & 0.31M & 61.76 & 86.31 & 88.64 & 92.52 & 90.16 & 78.56 & 93.54 & 90.88 & 85.30 \\ 
        Adapter\textsuperscript{P}\textsuperscript{$\dagger$} & 0.30M & 62.92 & 86.23 & 88.74 & 92.59 & 89.94 & 79.07 & 93.24 & 90.44 & 85.40 \\ 
        \midrule
        LoRA\textsubscript{ $r$=1} & 0.17M & 56.22 & 85.87 & 87.25 & 91.34 & 90.64 & 75.28 & 93.46 & 88.73 & 83.59 \\
        LoRA\textsubscript{ $r$=8} & 1.34M & 61.69 & 86.82 & 88.34 & 92.31 & 91.33 & 78.34 & 93.69 & 90.88 & 85.43 \\
        LoRA\textsubscript{ $r$=16} & 3.27M & 64.44 & 84.88 & 88.97 & 92.02 & \bf91.35 & 77.62 & 92.47 & \bf91.18 & 85.37 \\
        \midrule
        AdaLoRA\textsubscript{ $r$=1} & 0.67M & 57.86 & 87.21 & 88.24 & 92.46 & 89.91 & 76.17 & 93.69 & 89.99 & 84.44 \\
        AdaLoRA\textsubscript{ $r$=8} & 2.01M & 58.08 & 87.50 & 87.45 & 92.37 & 90.58 & 74.65 & 94.04 & 90.03 & 84.34 \\
        AdaLoRA\textsubscript{ $r$=16} & 4.02M & 59.35 & 87.67 & 88.73 & \bf92.64 & 90.79 & 77.26 & 93.23 & 90.26 & 84.99 \\
        \midrule
        ReLoRA\textsubscript{$1 \times 8$} & 0.17M & 59.91 & 85.61 & 86.11 & 89.13 & 87.20 & 82.54 & 93.44 & 89.20 & 84.14 \\
        \midrule \rowcolor{gray!10}
        \textit{\textbf{AROMA}} & 0.17M & \bf70.51 & 86.96 & \bf94.17 & 91.30 & 89.49 & \bf90.48 & \bf94.68 & 90.34 & \bf88.49 \\
        \bottomrule[1pt]
        \end{tabular}
    \caption{Comparative performance of different fine-tuning schemes for RoBERTa-base on GLUE benchmark. We report Matthew's correlation coefficient (MC) for CoLA, Pearson correlation coefficient (PC) for STS-B, and accuracy for all the remaining tasks. Higher is better for all metrics and the best results on each task are shown in \textbf{bold}. Results with "\textsuperscript{$\sharp$}" are retrieved from \citep{wang2024kasa}, and results with "\textsuperscript{$\dagger$}" are from \citep{mao2024dora}. Note that "\#Param" reflects the initial phase, and AROMA's \#Param gradually descends to zero (see Figure \ref{subfig param vs step}).}
    \vspace{-3mm}
    \label{table: glue}
\end{table*}

In this section, We fine-tune two LLMs of different sizes and architectures on two downstream tasks to evaluate the efficacy of AROMA. First, for natural language understanding (NLU) tasks, we fine-tune RoBERTa-base \citep{liu2019roberta} on the General Language Understanding Evaluation (GLUE) \citep{wang2018glue} benchmark. Second, for commonsense reasoning tasks, we fine-tune LLaMA3-8B \citep{llama3} on the Commonsense170K \citep{hu2023llm} dataset. 
NLU experiments are conducted on a single NVIDIA Tesla V100s-PCIE (32GB) GPU while the commonsense reasoning tasks are performed on two NVIDIA A100-SXM4 (80GB) GPUs. All the results reported in this section are averaged over multiple experiments with different random seeds.

\subsection{Baselines}\label{baselines}
Full fine-tuning and six PEFT methods serves as baselines, which are categorized into three groups:

\noindent \textbf{Adapter-based Methods.} 1) Adapter\textsuperscript{H} \citep{houlsby2019parameter}, which inserts lightweight adapter modules sequentially after transformer layers; and 2) Adapter\textsuperscript{P} \citep{pfeiffer2020adapterfusion}, which places adapters after feedforward network (FNN) and LayerNorm modules.

\noindent \textbf{LoRA-based Methods.} 1) LoRA;
2) AdaLoRA;
and 3) ReLoRA \citep{lialin2023relora}, which trains $K$ rank-$r$ matrices sequentially and merges them. While ReLoRA is designed for pretraining, it can be regarded as a reduced version of our method, where $T_{\mathrm{in}}$ and $T$ are fixed for all modules, and \eqref{check_in} and \eqref{check_out} are omitted. Therefore, we incorporate it to highlight the effectiveness of AriLoRA's adaptability and flexibility. 

\begin{figure*}[t]
  \centering
  \subfloat[Rank distribution of AdaLoRA]
  {
      \label{rank adalora mrpc}\includegraphics[width=0.5\linewidth]{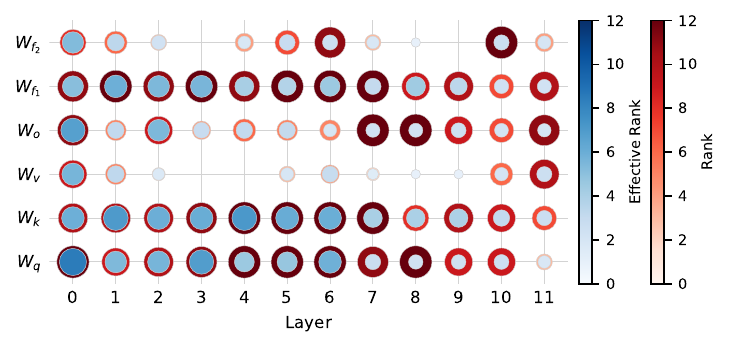}
  }
  \centering
  \subfloat[Rank distribution of \textit{\textbf{AROMA}}]
  {
      \label{rank AROMA mrpc}\includegraphics[width=0.5\linewidth]{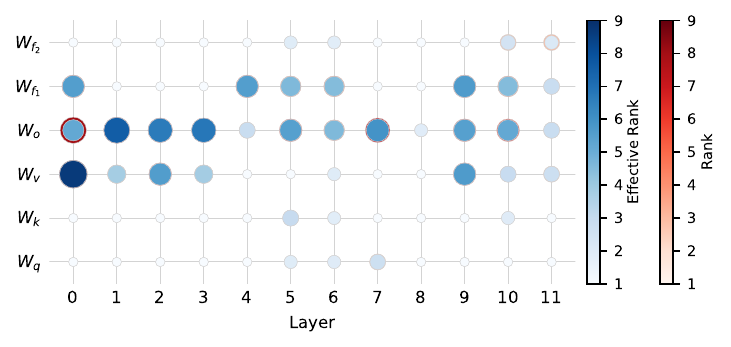}
  }
  \caption{Resultant rank and effective rank distributions for RoBERTa-base fine-tuned on MRPC task by AdaLoRA\textsubscript{$r$=8} and AROMA, respectively. The $x$-axis represents the hidden layer index, while the $y$-axis refers to the weight matrix fine-tuned in each layer. The total rank is described by the red outer circle, whereas the effective rank is indicated by the blue inner circle. Experiment on RTE task is provided in Appendix \ref{Apdx: RTE}.}
  \vspace{-3mm}
  \label{fig: rank distribution mrpc}
\end{figure*}

\noindent \textbf{Other Methods.} 1) Full fine-tuning, which updates all of the model's parameters; and 2) BitFit \citep{zaken2021bitfit}, which fine-tunes only the bias terms of a pretrained model. 

\subsection{Natural Language Understanding}
We first evaluate AROMA on NLU tasks. The model and datasets, training details are reported, followed by the results and analyses.

\noindent \textbf{Model and Datasets.} 
RoBERTa-base (125M) \citep{liu2019roberta} enhances BERT \citep{devlin2019bert} by utilizing larger batches, more data, and longer sequences, resulting in a stronger language understanding capability. Eight NLU tasks in GLUE (detailed in Appendix \ref{Apdx: glue}) are utilized to fine-tune RoBERTa-base, covering sentiment analysis, textual entailment, and semantic similarity.

\noindent \textbf{Training Details.}
To secure a fair comparison, we basically follow the implementation strategy in \citep{zhang2023adalora}. For each task in GLUE, we conduct a grid search for optimal hyperparameters, including the learning rate $lr \in$ [1E-4, 2E-4, 5E-4, 7E-4], inner tolerance $\varepsilon_\mathrm{in}$=0.1, and outer tolerance $\varepsilon_\mathrm{out} \in$ [1E-3, 5E-3, 6E-3]. We apply AROMA to all weight matrices, i.e., $\bm{W}_{q}$, $\bm{W}_{k}$, $\bm{W}_{v}$,$\bm{W}_{o}$, $\bm{W}_{f_1}$, and $\bm{W}_{f_2}$.

LoRA and AdaLoRA are conducted using the standard HuggingFace PEFT library, and the hyperparameters are set as suggested in their original papers. We consider the rank of LoRA and the target rank of AdaLoRA across $\{1,8,16\}$. The corresponding AdaLoRA's initial rank is set to $\{4, 12, 24\}$. For ReLoRA, rank $r=1$ is assigned to each LoRA to match the parameter budget. Detailed hyperparameter settings for each baseline are found in Appendix \ref{Apx Param1}.
\begin{table*}[t]
    \centering
    \fontsize{9}{13}\selectfont 
    \setlength{\tabcolsep}{4.0pt} 
        \begin{tabular}{c|c|cccccccc|c}
        \toprule[1pt] \rowcolor{pink!20}
        \bf Scheme & \bf \#Param & \textbf{\texttt{ARC-E}} & \textbf{\texttt{OBQA}} & \textbf{\texttt{SIQA}} & \textbf{\texttt{ARC-C}} & \textbf{\texttt{WinoG}} & \textbf{\texttt{PIQA}} &  \textbf{\texttt{BoolQ}} & \textbf{\texttt{HellaS}} & \textbf{Avg} \\ 
        \midrule
        ChatGPT\textsuperscript{$\lozenge$} & - & \bf89.7 & 74.8 & 68.5 & \bf79.9 & 66.1 & 85.4 & \underline{73.1} & 78.5 & 77.0 \\
        \midrule
        LoRA\textsubscript{ $r$=1} & 1.77M & 89.04 & 82.80 & 77.33 & 76.71 & 81.93 & {86.40} & 70.40 & 93.06 & 82.21 \\
        LoRA\textsubscript{ $r$=8} & 14.16M & 88.55 & 82.80 & 78.15 & 77.13 & \bf85.71 & 86.13 & 68.44 & {93.55} & 82.56 \\
        LoRA\textsubscript{ $r$=16} & 28.31M & 88.01 & {83.10} & \underline{79.53} & 75.34 & 83.82 & 85.74 & {72.35} & 93.45 & {82.67} \\ 
        \midrule
        AdaLoRA\textsubscript{ $r$=1} & 7.08M & 87.58 & 71.00 & 71.14 & 71.16 & 70.09 & 83.95 & 62.17 & 67.33 & 73.05 \\
        AdaLoRA\textsubscript{ $r$=8} & 21.23M & 88.30 & 76.60 & 71.24 & 71.33 & 72.45 & 83.51 & 65.57 & 82.94 & 76.49 \\
        AdaLoRA\textsubscript{ $r$=16} & 42.47M & 88.47 & 75.20 & 71.14 & 72.70 & 71.90 & 84.17 & 62.69 & 84.13 & 76.30 \\ 
        \midrule \rowcolor{gray!10}
        \textit{\textbf{AROMA}}\textsubscript{ $r$=1} & 1.77M & 89.31 & \underline{83.70} & 79.12 & 78.50 & 81.85 & \bf87.43 & 71.16 & \underline{93.79} & \underline{83.11} \\ \rowcolor{gray!10}
        \textit{\textbf{AROMA}}\textsubscript{ $r$=8} & 14.16M & \underline{89.48} & \bf84.79 & \bf{79.62} & \underline{78.76} & \underline{83.98} & \underline{87.22} & \bf73.74 & \bf94.36 & \bf83.85 \\
        \bottomrule[1pt]
        \end{tabular}
    \caption{Comparative performance of different fine-tuning schemes for LLaMA3-8B on Commonsense170K dataset. We report accuracy for all tasks. 
    Results with "\textsuperscript{$\lozenge$}" are retrieved from \citep{liu2024dora}. 
    Note that "\#Param" reflects the number of initial trainable parameters, and AROMA's average \#Param is even less.}
    \vspace{-2mm}
    \label{table: commonsense}
\end{table*}

\noindent \textbf{Results and Analyses.}
Table \ref{table: glue} presents the performance of AROMA alongside its counterparts, where "\#Param" refers to the number of initial trainable parameters. It is shown that both AdaLoRA and LoRA are sensitive to the rank parameter, whereas AROMA operates independently of it. AROMA achieves the highest average performance. In term of specific tasks, it surpasses other baselines on CoLA, MRPC, RTE, and SST-2, while yields comparable results on the remaining tasks. This is achieved with only 0.014\% (approximately 0.17M out of 125.0M) of the trainable parameters required for full fine-tuning. In comparison to ReLoRA, a reduced version of AROMA without rank adaptability, our method demonstrates superiority on all tasks, showcasing the latter effectiveness. Particularly, AROMA shows a significant advantage in CoLA, MRPC, and RTE tasks. We will further explore MRPC and RTE to analyze the reasons behind AROMA's outstanding performance.

We plot the rank distributions for AdaLoRA and AROMA in Figs. \ref{fig: rank distribution mrpc} and \ref{fig: rank distribution rte}, where the rank is a combination of effective rank \cite{roy2007effective} and non-effective rank. The former measures the effective dimensionality of a matrix, while the latter corresponds to dimensions with negligible contribution. Detailed description of effective rank are provided in Appendix \ref{Apdx: eff rank}. It is observed that different weight matrices exhibit distinct rank characteristics, and AdaLoRA has a larger average rank than AROMA. Furthermore, the rank distribution for AROMA is concentrated in the shallower layers, $\bm{W}_v$ and $\bm{W}_o$ for both MRPC and RTE tasks. In terms of effective rank, it is found that LoRA exhibits a low effective rank, just a quarter of the adapter rank \citep{shuttleworth2024lora,biderman2024lora,he2025rasa}. For AdaLoRA, we see that only about half of its rank is effective (50.4\% for MRPC, 49.2\% for RTE), whereas AROMA exhibits an exceptionally high effective rank ratio (96.3\% for MRPC and 91.7\% for RTE). 

Moreover, Figure \ref{fig: training steps} depicts the number of trainable parameters, total rank, ranks of specific layers and accuracy versus training step for RoBERTa-base on MRPC task. We select "layer.0.attention.output.dense" and "layer.9.attention.self.value" as illustration. It is evident that LoRA\textsubscript{$r$=8}, AdaLoRA\textsubscript{$r$=8} and AROMA exhibit consistent, decreasing and growing rank behaviors, respectively. We notice that LoRA maintains nearly 1.3M trainable parameters, with a stable total rank and specific rank throughout, as it fixes the same rank for all weight matrices. AdaLoRA, on the other hand, progressively decreases the total rank and shows a fluctuating but generally declining specific rank, starting with 2.0M trainable parameters and averaging 1.62M. In contrast, AROMA necessitates only 0.17M trainable parameters initially, with an average of 0.08M. Remarkably, AROMA attains the highest accuracy among the three methods.







\subsection{Commonsense Reasoning}

In this section, we assess AROMA in handling a larger model and a more complex task.

\noindent \textbf{Model and Datasets.}
Following \citep{wang2024kasa}, we fine-tune LLaMA3-8B \citep{llama3} on the Commonsense170K dataset, which is a mixture of eight commonsense reasoning benchmarks (details provided in Appendix \ref{Apdx: CR}). LLaMA3-8B model, developed by Meta, is designed for various NLP tasks, offering improved performance and efficiency over its predecessors.

\noindent \textbf{Training Details.}
Apart from AROMA under the previous setting, we additionally increase the rank of each LoRA update to 8 to accommodate this complex task. We apply AROMA to three weight matrices in the self-attention layer: $\bm{W}_{q}$, $\bm{W}_{k}$, $\bm{W}_{v}$, and two in the FFN: $\bm{W}_{up}$, and $\bm{W}_{down}$. After fine-tuning, the resultant model is evaluated on each of the eight benchmarks in terms of accuracy. Detailed hyperparameter settings are found in Appendix \ref{Apx Param2}. 

\noindent \textbf{Results.}
Table \ref{table: commonsense} shows the comparative performance between AROMA and its counterparts, where ChatGPT \citep{wei2022chain} is also included for reference. Notably, AROMA\textsubscript{$r$=1} and AROMA\textsubscript{$r$=8} rank in the top two in terms of average accuracy. Specifically, AROMA\textsubscript{$r$=1} achieves this with approximately 0.02\% of the original model's parameters, 6\% of LoRA\textsubscript{$r$=8}'s and 3\% of AdaLoRA\textsubscript{$r$=8}'s. AROMA\textsubscript{$r$=8} outpaces other baselines on three benchmarks and achieves second-best results on the remaining ones. These results validates the efficacy of our method.


\section{Further Discussions}

\subsection{Ablation Study}
We carry out ablation study on a crucial component of AROMA: \emph{Reset}, i.e., randomly pruning 99.9\% of the optimizer states after training a rank-one update, to validate its effectiveness on performance. We fine-tune RoBERTa-base on MRPC task using AROMA with and without \emph{Reset}, respectively, with all other conditions remain unchanged. We average the results over 5 experiments with different seeds, and report the average rank and effective rank across all layers as well as accuracy. 
\begin{table}[ht]
    \centering
    \fontsize{9}{13}\selectfont 
    \setlength{\tabcolsep}{2.5pt} 
        \begin{tabular}{c | c c c | c c c}
        \toprule[1pt] \rowcolor{cyan!10}
        & \multicolumn{3}{c|}{\textbf{\texttt{MRPC}}} & \multicolumn{3}{c}{\textbf{\texttt{RTE}}} \\ \rowcolor{cyan!10}
        \multirow{-2}{*}{\bf Scheme} & Avg $r$ & Eff $r$ & Acc & Avg $r$ & Eff $r$ & Acc \\
        \midrule
        \textit{\textbf{AROMA}}\textsubscript{w/o \emph{Reset}} & 1.43 & 1.39 & 83.33 & 1.42 & 1.30 & 70.48 \\ \rowcolor{gray!10}
        \textit{\textbf{AROMA}}\textsubscript{w/ \emph{Reset}} & 2.78 & 2.68 & \bf94.17 & 3.42 & 3.14 & \bf90.48 \\
        \bottomrule[1pt]
        \end{tabular}
    \caption{Comparison of AROMA with and without optimizer \emph{Reset} for RoBERTa-base on MRPC task. "Avg $r$" and "Eff $r$" denote average rank and average effective rank, respectively.}
    \vspace{-2mm}
    \label{table: ablation}
\end{table}

As seen in Table \ref{table: ablation}, AROMA with the \emph{Reset} mechanism demonstrates a larger rank than AROMA\textsubscript{w/o \emph{Reset}} and achieves substantially higher accuracy. This suggests that \emph{Reset} is beneficial. We interpret this as the optimizer reset allowing the new rank-one matrix to be computed from scratch, rather than relying on the previously computed rank-one matrix. This approach gives the new rank-one matrix a greater chance to explore new subspaces and learn more information. Supplementary experiment on cosine similarity in Appendix \ref{Apdx: cosine} further underscores the importance of the \emph{Reset} mechanism.

\subsection{Time Efficiency}\label{efficiency}
In this subsection, we compare the efficiency of AROMA with LoRA and AdaLoRA. We unify the three methods by configuring their batch size of 64 and maximum sequence length of 256, and compute the average training time per epoch across six tasks in the GLUE benchmark on a single NVIDIA Tesla V100s-PCIE (32GB) GPU. The results are reported in Table \ref{table: efficiency} and we see that AROMA demonstrates significant efficiency advantages in five tasks, while being comparable to LoRA in the remaining task, RTE. Particularly, its average time per epoch is 76.1\% of LoRA's and 28.5\% of AdaLoRA's. This superiority can be attributed to the rank-one training and unnecessity of SVD computation.


\begin{table}[htbp]
    \centering
    \fontsize{9}{13}\selectfont 
    \setlength{\tabcolsep}{3.5pt} 
        \begin{tabular}{c | c c c}
        \toprule[1pt] \rowcolor{cyan!10}
        \textbf{Task} & {LoRA} & {AdaLoRA} & \textit{\textbf{AROMA}}\\
        \midrule
        \textbf{\texttt{CoLA}} & 44.37 & 107.74 & \cellcolor{gray!10} \bf12.43 \\
        \textbf{\texttt{MRPC}} & 17.84 & 45.57 & \cellcolor{gray!10} \bf13.21 \\
        \textbf{\texttt{QNLI}} & 557.98 & 1547.82 & \cellcolor{gray!10} \bf542.72 \\
        \textbf{\texttt{RTE}} & \bf15.13 & 31.46 & \cellcolor{gray!10} 20.14 \\
        \textbf{\texttt{SST-2}} & 339.58 & 873.30 & \cellcolor{gray!10} \bf 153.47 \\
        \textbf{\texttt{STS-B}} & 30.04 & 73.13 & \cellcolor{gray!10} \bf22.42 \\
        \midrule
        \textbf{Avg} & 167.50 & 446.50 & \cellcolor{gray!10} \bf127.40 \\
        \bottomrule[1pt]
        \end{tabular}
    \caption{Per-epoch time comparison for RoBERTa-base on GLUE.}
    \vspace{-2mm}
    \label{table: efficiency}
\end{table}

\section{Related Work}
PEFT emerges as a crucial approach for adapting LLMs to downstream tasks while minimizing computational and storage requirements. We categorize existing PEFT methods into three key paradigms \citep{han2024parameter} as follows:

\noindent \textbf{Additive PEFT Methods} incorporate auxiliary trainable modules within transformer architectures. Serial adapter \citep{houlsby2019parameter} introduces dual adapter modules positioned after self-attention and FFN layers, while \citep{pfeiffer2020adapterfusion} optimizes computational efficiency by inserting adapters exclusively after "Add \& Norm" layers. 
Prompt-based techniques constitute another significant branch of additive PEFT. Approaches such as prefix-tuning \citep{liliang2021prefix, li2023prefix, zhang2023towards}, p-tuning \citep{liu2024gpt}, and prompt-tuning \citep{lester2021power} augment inputs or intermediate representations with trainable vectors, demonstrating particular efficacy for generative tasks and few-shot learning scenarios.

\noindent \textbf{Selective PEFT Methods} strategically identify and modify only the most critical subset of model parameters. BitFit \citep{zaken2021bitfit} achieves remarkable efficiency by exclusively fine-tuning bias terms while maintaining all other parameters frozen. Diff pruning \citep{guo2020parameter} learns sparse parameter differences from pretrained weights, focusing on task-specific components. FishMask \citep{sung2021training} leverages Fisher information to identify and update the most influential parameters for specific tasks. 

\noindent \textbf{Reparameterized PEFT Methods} transform the parameter space to facilitate efficient updates without direct modification of original weights. (IA)\textsuperscript{3} \citep{liu2022few} and SSF \citep{lian2022scaling} introduce learnable vectors that modulate activations in self-attention and FFN with low parameter overhead. LoRA \citep{hu2022lora} decomposes weight updates into low-rank matrix products, significantly reducing trainable parameters while preserving performance. AdaLoRA \citep{zhang2023adalora} enhances flexibility through SVD-like decomposition for dynamic rank allocation. DoRA \citep{liu2024dora} decomposes the weight into magnitude and directional components. NOLA \citep{koohpayegani2024nola} and VeRA \citep{kopiczko2024vera} represent weight matrices as linear combinations of fixed random bases,  optimizing only the mixture coefficients. HydraLoRA \citep{tian2024hydralora} maintains fixed LoRA $\bm{A}$ matrix while training multiple $\bm{B}$ matrices to accommodate multi-domain tasks. LoRA and its variants achieve state-of-the-art parameter efficiency, making them the most widely used PEFT approaches.

\section{Conclusion}
In this work, we propose \textbf{A}utonomous \textbf{R}ank-\textbf{O}ne \textbf{M}atrix \textbf{A}daptation (AROMA) for parameter-efficient fine-tuning. Unlike the existing adaptive rank adjustment method, AdaLoRA, which truncates singular values with low importance scores and requires both initial and target rank budgets, AROMA employs a rank-growing approach that autonomously constructs layer-specific updates with very few trainable parameters that gradually diminish to zero. We design a dual-loop architecture, featuring an inner loop that exploits each rank-one subspace to learn a LoRA update with the corresponding stopping criterion, while the outer loop determines the number of subspaces, namely, the optimal rank, guided by another stopping criterion. The learned rank-one components are merged and frozen, allowing only one rank-one LoRA to be trained at a time, thereby ensuring high parameter efficiency. Additionally, optimizer states are periodically reset to maintain subspace independence. Experimental results for NLU and commonsense reasoning tasks highlight AROMA's superiority in terms of accuracy and efficiency. In future research, we plan to apply AROMA to more tasks and explore its potential in continual learning settings, where models are sequentially adapted to multiple tasks.

\section*{Limitations}
Despite achieving promising results on NLU and commonsense reasoning benchmarks, our approach has several challenges to be tackled. It has yet to be tested in multimodal applications, a crucial area as multimodal models continue to gain prominence. Furthermore, we have not validated its scalability for extremely LLMs exceeding 100 billion parameters, where the dynamics of rank allocation may differ significantly. Future work should address these issues and explore the method's applicability across a broader range of tasks.

\section*{Acknowledgments}

This document has been adapted
by Steven Bethard, Ryan Cotterell and Rui Yan
from the instructions for earlier ACL and NAACL proceedings, including those for
ACL 2019 by Douwe Kiela and Ivan Vuli\'{c},
NAACL 2019 by Stephanie Lukin and Alla Roskovskaya,
ACL 2018 by Shay Cohen, Kevin Gimpel, and Wei Lu,
NAACL 2018 by Margaret Mitchell and Stephanie Lukin,
Bib\TeX{} suggestions for (NA)ACL 2017/2018 from Jason Eisner,
ACL 2017 by Dan Gildea and Min-Yen Kan,
NAACL 2017 by Margaret Mitchell,
ACL 2012 by Maggie Li and Michael White,
ACL 2010 by Jing-Shin Chang and Philipp Koehn,
ACL 2008 by Johanna D. Moore, Simone Teufel, James Allan, and Sadaoki Furui,
ACL 2005 by Hwee Tou Ng and Kemal Oflazer,
ACL 2002 by Eugene Charniak and Dekang Lin,
and earlier ACL and EACL formats written by several people, including
John Chen, Henry S. Thompson and Donald Walker.
Additional elements were taken from the formatting instructions of the \emph{International Joint Conference on Artificial Intelligence} and the \emph{Conference on Computer Vision and Pattern Recognition}.

\bibliography{acl_latex}

\clearpage
\appendix

\section{Algorithm of AROMA}\label{Apdx: alg1}
We present the details of AROMA in Algorithm \ref{Alg1}. 

\section{Time Complexity}\label{Apdx: complexity}
We first analyze the per-step complexity to calculate $\varDelta \bm{W}$ of dimensions $m\times n$. In the forward pass, considering $\bm{B}\in\mathbb{R}^{m\times r}$, $\bm{A}\in\mathbb{R}^{r\times n}$, and $\bm{x}\in\mathbb{R}^{n}$. LoRA costs $\mathcal{O}((m+n)r)$ time. AdaLoRA calculates $\bm{P\varLambda Qx}$, hence its complexity is $\mathcal{O}((m+n+\tilde{r})\tilde{r})=\mathcal{O}((m+n)\tilde{r})$, where $\tilde{r}$ is the current rank. AROMA computes $\bm{B_pA_px}$ with $p$ being the current outer step, which requires $\mathcal{O}\left(\left(m+n\right)p\right)$ time. Since LoRA has a consistent rank, AdaLoRA decreases rank, while AROMA increases rank, typically we have $\tilde{r}\geq r \geq p$, which leads to $\mathcal{O}_{\text{per-step}}^{\text{AdaLoRA}} > \mathcal{O}_{\text{per-step}}^{\text{LoRA}} \geq \mathcal{O}_{\text{per-step}}^{\text{AROMA}}$.

Based on this, we discuss the overall complexity. Given $T$ as the total training steps, LoRA consumes $\mathcal{O}\left((m+n)rT\right)$ time. For AdaLoRA, we roughly denote its average rank as $\frac{r_i+r_f}{2}$ with $r_i$ and $r_f$ being the initial average rank and the target average rank, respectively, then its overall complexity is $\mathcal{O}\left((m+n)\frac{r_i+r_f}{2}T\right)$. For AROMA, supposing that each inner loop has $T_{\mathrm{in}}$ steps for simplicity, and there are $P$ outer steps, i.e., $T=P\cdot T_{\mathrm{in}}$, the overall complexity is $\mathcal{O}\left((m+n)T_{\mathrm{in}}\sum_{p=1}^{P}{p}\right)=\mathcal{O}\left( (m+n)\frac{1+P}{2}T \right)$. Typically, we have $\mathcal{O}_{\text{overall}}^{\text{AdaLoRA}} > \mathcal{O}_{\text{overall}}^{\text{LoRA}} \geq \mathcal{O}_{\text{overall}}^{\text{AROMA}}$. The above claims are listed in Table \ref{table: detailed complexity} and are experimentally validated in Section \ref{efficiency}.
\begin{table}[ht]
    \centering
    \fontsize{8}{11}\selectfont 
    \setlength{\tabcolsep}{1pt} 
        \begin{tabular}{c | c c c}
        \toprule[1pt] \rowcolor{cyan!10}
        \bf Scheme & LoRA & AdaLoRA & \textit{\textbf{AROMA}} \\
        \midrule
        \tiny{\makecell{\bf Per-step \\ \bf Complexity}} & $\mathcal{O}((m+n)r)$ & {$\mathcal{O}((m+n)\tilde{r})$} & \cellcolor{gray!10} $\mathcal{O}((m+n)p)$ \\
        \midrule
        \tiny{\makecell{\bf Overall \\ \bf Complexity}} & \tiny$\mathcal{O}((m+n)r T)$ & \tiny$\displaystyle\mathcal{O}(\frac{r_i+r_f}{2}(m+n)T)$ & \cellcolor{gray!10} \tiny$\mathcal{O}\left( (m+n)T\frac{1+P}{2} \right)$ \\
        \bottomrule[1pt]
        \end{tabular}
    \caption{Complexity comparison}
    \label{table: detailed complexity}
\end{table}

\section{Effective Rank}\label{Apdx: eff rank}
In data representation, effective rank \citep{roy2007effective} reflects the number of truly meaningful independent feature dimensions in a matrix, whose definition is given as follows. Consider a $m\times n$ matrix $\bm{W}$ with singular values:
\begin{equation}
    \sigma_1 \geq \sigma_2 \geq \cdots \geq \sigma_K \geq 0
\end{equation}
where $K=\min \left\{ m,n \right\} $. Given $p_k=\frac{\sigma_k}{\sum_{k=1}^K{|\sigma_k|}}$, the effective rank is defined as: 
\begin{equation}
    \mathrm{erank}=\exp\left\{H(p_1,p_2,\cdots,p_K)\right\}
\end{equation}
where $H(p_1,p_2,\cdots,p_K)$ is the Shannon entropy:
\begin{equation}
    H(p_1,p_2,\cdots,p_K)=-\sum_{k=1}^{K}{p_k\log p_k}
\end{equation}
Effective rank is smaller than full rank as it ignores dimensions with minimal contributions.

In neural network weight matrices, effective rank indicates the number of effective feature transformations learned by that layer. Low effective rank proportion suggests redundancy or underutilized parameters \citep{shuttleworth2024lora}.

\section{Rank Distribution for RTE Task}\label{Apdx: RTE}
\begin{figure*}
  \centering
  \subfloat[Rank distribution of AdaLoRA]
  {
      \label{rank adalora rte}\includegraphics[width=0.5\linewidth]{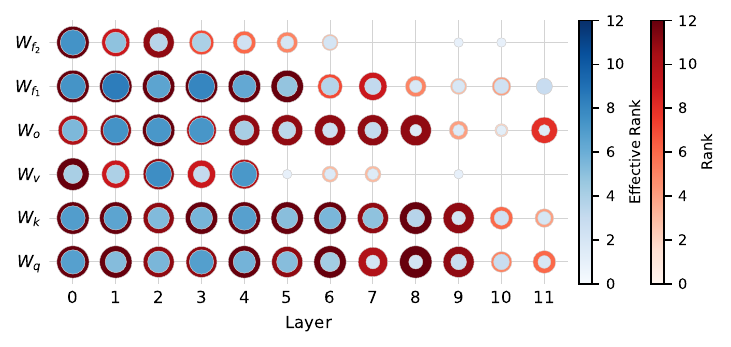}
  }
  \centering
  \subfloat[Rank distribution of \textit{\textbf{AROMA}}]
  {
      \label{rank AROMA rte}\includegraphics[width=0.5\linewidth]{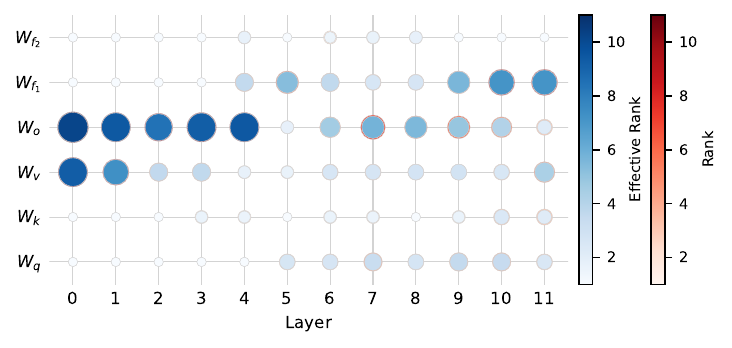}
  }
  \caption{Resultant rank and effective rank distributions for RoBERTa-base fine-tuned on RTE task by AdaLoRA\textsubscript{$r$=8} and AROMA, respectively. The $x$-axis represents the hidden layer index, while the $y$-axis refers to the weight matrix fine-tuned in each layer. The total rank is described by the red outer circle, whereas the effective rank is indicated by the blue inner circle.}    
  \label{fig: rank distribution rte}
\end{figure*}

Figure \ref{fig: rank distribution rte} shows the rank distributions for AdaLoRA and AROMA on RTE task, and we observe a similar phenomenon to that of Figure \ref{fig: rank distribution mrpc}.

\section{Cosine Similarity}\label{Apdx: cosine}
\begin{figure}[ht]
    \centering
    \includegraphics[width=\linewidth]{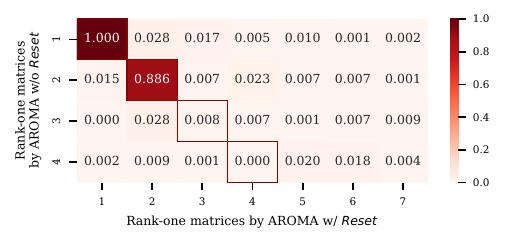}
    \caption{Cosine similarity between AROMA\textsubscript{w/o \emph{Reset}} and AROMA\textsubscript{w/ \emph{Reset}} for layer.10.attention.output.sense layer results for RoBERTa-base on MRPC task.}
    \label{fig:cosine similarity}
\end{figure}
Figure \ref{fig:cosine similarity} shows the cosine similarity between AROMA\textsubscript{w/o \emph{Reset}} and AROMA\textsubscript{w/ \emph{Reset}}, which we only focus on values on the diagonal. It reveals that their solutions are identical initially, but increasingly diverge with each subsequent \emph{Reset}. This finding further underscores the important role of the \emph{Reset} mechanism.

\section{Dataset Details}
\subsection{GLUE}\label{Apdx: glue}
GLUE \citep{wang2018glue} is a collection of nine NLU benchmarks designed to evaluate the performance of LLMs across multiple dimensions of linguistic competence. This work involves eight commonly used GLUE tasks: CoLA \citep{warstadt2019neural}, MNLI \citep{williams2017broad}, MRPC \citep{dolan2005automatically}, QNLI \citep{rajpurkar2016squad}, QQP \citep{wang2018glue}, RTE \citep{dagan2005pascal, haim2006second, giampiccolo2007third, bentivogli2009fifth}, SST-2 \citep{socher2013recursive}, STS-B \citep{wang2018glue}. Their details are listed in Table \ref{table: glue details}.
\begin{table}[htbp]
    \centering
    \fontsize{9}{13}\selectfont 
    \setlength{\tabcolsep}{4pt} 
        \begin{tabular}{c|ccccc}
        \toprule[1pt] \rowcolor{cyan!10}
        \bf Dataset & \bf\#Train & \bf\#Valid & \bf\#Test & \bf\#Label & \bf Metric \\
        \midrule \rowcolor{gray!10}
        \multicolumn{6}{c}{Single-Sentence Classification} \\
        \textbf{\texttt{CoLA}} & 8.5k & 1k & 1k & 2 & MC \\
        \textbf{\texttt{SST-2}} & 67k & 872 & 1.8k & 2 & Acc \\
        \midrule \rowcolor{gray!10}
        \multicolumn{6}{c}{Pairwise Text Classification} \\
        \textbf{\texttt{MNLI}} & 393k & 20k & 20k & 3 & Acc \\
        \textbf{\texttt{RTE}} & 2.5k & 277 & 3k & 2 & Acc \\
        \textbf{\texttt{QQP}} & 364k & 40k & 391k & 2 & Acc \\
        \textbf{\texttt{MRPC}} & 3.7k & 408 & 1.7k & 2 & Acc \\
        \textbf{\texttt{QNLI}} & 105k & 5.5k & 5.5k & 2 & Acc \\
        \midrule \rowcolor{gray!10}
        \multicolumn{6}{c}{Text Similarity} \\
        \textbf{\texttt{STS-B}} & 5.7k & 1.5k & 1.4k & 1 & PC \\
        \bottomrule[1pt]
        \end{tabular}
    \caption{Details of GLUE benchmark. "MC", "PC", and "Acc" represent Matthews correlation coefficient, Pearson correlation coefficient, and accuracy, respectively. "\#Train", "\#Valid", and "\#Test" refer to the number of training, validation, and testing examples, respectively. "\#Label" denotes the number of labels.}
    \label{table: glue details}
\end{table}

\subsection{Commonsense170K}\label{Apdx: CR}
Commonsense170K \citep{hu2023llm} is a comprehensive benchmark collection comprising approximately 170,000 training examples and 400 validation examples across eight diverse commonsense reasoning datasets: ARC-Easy and ARC-Challenge \citep{clark2018think}, OBQA \citep{mihaylov2018can}, SIQA \citep{sap2019socialiqa}, WinoGrande \citep{sakaguchi2021winogrande}, PIQA \citep{bisk2020piqa}, BoolQ \citep{clark2019boolq}; and HellaSwag \citep{zellers2019hellaswag}. This consolidated benchmark evaluates LLMs' capabilities across multiple dimensions of commonsense knowledge, including conceptual reasoning, physical understanding, social intelligence, causal reasoning, coreference resolution, and scientific knowledge.


\section{Hyperparameter Settings}
\subsection{NLU Task}\label{Apx Param1}
Hyperparameter setup for NLU task can be found in Table \ref{table: glue param}, where we follow the suggested setting for LoRA and AdaLoRA, and meticulously tune for AROMA, including the learning rate $lr \in$ [1E-4, 2E-4, 5E-4, 7E-4], inner tolerance $\varepsilon_\mathrm{in} \in$ [0.05, 0.1], and outer tolerance $\varepsilon_\mathrm{out} \in$ [1E-3, 5E-3, 6E-3]. Initial warmup is 100 and subsequent warmup is 50 for all tasks, except CoLA which uses 500 and 100 respectively. We use publicly available implementation (\url{https://github.com/Guitaricet/relora}) to run ReLoRA.

\subsection{Commonsense Reasoning Task}\label{Apx Param2}
Hyperparameter setup for commonsense reasoning task can be found in Table \ref{table: CR param}.
\begin{table}[htbp]
    \centering
    \fontsize{9}{13}\selectfont 
        \begin{tabular}{c|c|c}
        \toprule[1pt] \rowcolor{pink!20}
        \bf Scheme & \bf Hyperparameter & \bf Value \\
        \midrule
        \multirow{12}{*}{\textit{\textbf{AROMA}}\textsubscript{$r$=1}} & 
        $r$ & 1 \\
        & $\alpha$ & 2 \\
        & Max Seq. Len. & 256 \\
        & Batch Size & 32 \\
        & Epoch & 20 \\
        & Learning Rate & 1E-4 \\
        & $T$ & 100,000 \\
        & $T_{\mathrm{in}}$ & 1000 \\
        & $\varDelta T_{\mathrm{in}}$ & 10 \\
        & $\varepsilon_\mathrm{in}$ & 0.1 \\
        & $\varepsilon_\mathrm{out}$ & 1E-3 \\
        & Eval Batch Size & 8 \\
        \midrule
        \multirow{12}{*}{\textit{\textbf{AROMA}}\textsubscript{$r$=8}} & 
        $r$ & 8 \\
        & $\alpha$ & 16 \\
        & Max Seq. Len. & 256 \\
        & Batch Size & 32 \\
        & Epoch & 15 \\
        & Learning Rate & 1E-4 \\
        & $T$ & 80,000 \\
        & $T_{\mathrm{in}}$ & 2000 \\
        & $\varDelta T_{\mathrm{in}}$ & 10 \\
        & $\varepsilon_\mathrm{in}$ & 0.1 \\
        & $\varepsilon_\mathrm{out}$ & 1E-2 \\
        & Eval Batch Size & 8 \\
        \bottomrule[1pt]
        \end{tabular}
    \caption{Hyperparameter setup for LLaMA3-8B on Commonsense170k}
    \label{table: CR param}
\end{table}

\begin{algorithm*}[t]
\caption{\textit{\textbf{AROMA}}}
\label{Alg1}
\DontPrintSemicolon
\SetKwFor{For}{for}{}{endfor}  
\vspace{0.5ex}
    \KwIn{Inner and outer tolerances $\varepsilon_\mathrm{in}$ and $\varepsilon_\mathrm{out}$, maximum inner training steps $T_{\mathrm{in}}$, inner checking interval $\varDelta T_{\mathrm{in}}$, maximum total training steps $T$.}
    \For{each module in parallel}{
        {\bf Initialize:} $\bm{b}_1^{(0)} \gets \bm{0}$; $\bm{a}_1^{(0)} \gets \mathrm{Kaiming\_init}$.\;
        Freeze $\bm{W}_0$. \;
        \For{$p=1,2, \cdots$ {\bf do} \, \, \, \, \, \, \, \, \, \, \, \textup{\texttt{// OUTER LOOP}}}{
            \For{$t=1,2,\cdots, T_{\mathrm{in}}$ {\bf do} \, \, \textup{\texttt{// INNER LOOP}}}{
                Update $\bm{b}_p^{(t)}$, $\bm{a}_p^{(t)}$. \;
                \If{$\mathrm{MOD}(t,\varDelta T_{\mathrm{in}})=0$}{
                    inner\_converged = True, {\bf if} $\frac{\left\| \boldsymbol{b}_p^{(t)}\boldsymbol{a}_p^{(t)} \right\| _F - \left\| \boldsymbol{b}_p^{(t-\varDelta T_{\mathrm{in}})}\boldsymbol{a}_p^{(t-\varDelta T_{\mathrm{in}})} \right\| _F}{\left\| \boldsymbol{b}_p^{(t-\varDelta T_{\mathrm{in}})}\boldsymbol{a}_p^{(t-\varDelta T_{\mathrm{in}})} \right\| _F}<\varepsilon_\mathrm{in}$. \texttt{// CHECK}\;
                    {\bf Break} the inner loop, {\bf if } all modules are inner\_converged. \;
                }
            }
            outer\_converged = True, {\bf if} $ \frac{\left\| \alpha \boldsymbol{b}_p\boldsymbol{a}_p \right\| _F}{\left\| \boldsymbol{W}_0+\alpha \boldsymbol{B}_{p-1}\boldsymbol{A}_{p-1} \right\| _F}<\varepsilon_\mathrm{out}$. \texttt{// CHECK}\; 
            {\bf Break} the outer loop, {\bf if } outer\_converged. \;
            $\varDelta \bm{W} = \varDelta \bm{W} + \bm{b}_p^{(t)} \bm{a}_p^{(t)} $. \, \, \, \, \, \, \, \, \, \, \, \, \, \, \, \, \, \, \, \, \, \, \, \, \, \texttt{// MERGE}\;
            $\bm{b}_{p+1}^{(0)} \gets \bm{0}$; $\bm{a}_{p+1}^{(0)} \gets \mathrm{Kaiming\_init}$. \, \, \, \, \, \, \, \, \, \, \, \, \, \, \, \, \, \texttt{// REINIT}\;
            Reset optimizer states \& learning rate warmup. \, \, \, \, \, \, \, \texttt{// RESET}\;
        }
        {\bf Finish}, {\bf if } all modules are outer\_converged {\bf or } reach $T$. \;
    }
    \KwOut{$\varDelta \bm{W}$.}
\end{algorithm*}

\begin{table*}
    \centering
    \fontsize{9}{13}\selectfont 
        \begin{tabular}{c|c|cccccccc}
        \toprule[1pt] \rowcolor{cyan!10}
        \bf Scheme & \bf Hyperparameter & \textbf{\texttt{CoLA}} & \textbf{\texttt{MNLI}} & \textbf{\texttt{MRPC}} & \textbf{\texttt{QNLI}} & \textbf{\texttt{QQP}} & \textbf{\texttt{RTE}} & \textbf{\texttt{SST-2}} & \textbf{\texttt{STS-B}} \\
        \midrule
        \multirow{6}{*}{LoRA} & Max Seq. Len. & \multicolumn{8}{c}{128} \\
        & Batch Size & \multicolumn{8}{c}{64} \\
        & Epoch & 30 & 30 & 30 & 25 & 25 & 50 & 60 & 40 \\
        & Learning Rate & 4E-4 & 5E-4 & 4E-4 & 4E-4 & 4E-4 & 5E-4 & 5E-4 & 4E-4 \\
        & $r$ & \multicolumn{8}{c}{8} \\
        & $\alpha$ & \multicolumn{8}{c}{16} \\
        \midrule
        \multirow{12}{*}{AdaLoRA} & Max Seq. Len. & \multicolumn{8}{c}{128} \\
        & Batch Size & \multicolumn{8}{c}{32} \\
        & Epoch & 25 & 7 & 30 & 5 & 5 & 52 & 24 & 26 \\
        & Learning Rate & 5E-4 & 5E-4 & 1E-3 & 1.2E-3 & 5E-4 & 1.2E-3 & 8E-4 & 2.2E-3 \\
        & $r_i$ & \multicolumn{8}{c}{12} \\
        & $r_f$ & \multicolumn{8}{c}{8} \\
        & $\gamma$ & 0.5 & 0.1 & 0.1 & 0.1 & 0.1 & 0.3 & 0.1 & 0.1 \\
        & $T$ & 6500 & 85000 & 3000 & 15000 & 55000 & 4000 & 50000 & 4500 \\
        & $t_i$ & 800 & 8000 & 600 & 2000 & 8000 & 600 & 6000 & 800 \\
        & $\varDelta_T$ & 10 & 100 & 1 & 100 & 100 & 1 & 100 & 10 \\
        & $t_f$ &3500 & 50000 & 1800 & 8000 & 25000 & 1800 & 22000 & 2000 \\
        & $\alpha$ & \multicolumn{8}{c}{32} \\
        \midrule \rowcolor{gray!10}
        & Max Seq. Len. & \multicolumn{8}{c}{256} \\ \rowcolor{gray!10}
        & Batch Size & 32 & 32 & 64 & 32 & 64 & 64 & 64 & 32 \\ \rowcolor{gray!10}
        & Epoch & 130 & 10 & 52 & 10 & 10 & 62 & 40 & 50 \\ \rowcolor{gray!10}
        & Learning Rate & 2E-4 & 7E-4 & 1E-4 & 2E-4 & 4E-4 & 1E-4 & 5E-4 & 5E-4 \\ \rowcolor{gray!10}
        & $T$ & 35000 & 85000 & 3000 & 30000 & 55000 & 2400 & 40000 & 10000 \\ \rowcolor{gray!10}
        & $T_{\mathrm{in}}$ & 5000 & 5000 & 200 & 2000 & 55000 & 200 & 2500 & 1000 \\ \rowcolor{gray!10}
        & $\varDelta T_{\mathrm{in}}$ & \multicolumn{8}{c}{10} \\ \rowcolor{gray!10}
        & $\varepsilon_\mathrm{in}$ & \multicolumn{8}{c}{0.1} \\ \rowcolor{gray!10}
        & $\varepsilon_\mathrm{out}$ & 2E-2 & 5E-3 & 5E-3 & 5E-3 & 1E-3 & 6E-3 & 5E-3 & 5E-3 \\ \rowcolor{gray!10}
        \multirow{-10}{*}{\textit{\textbf{AROMA}}} & $\alpha$ & \multicolumn{8}{c}{4} \\
        \bottomrule[1pt]
        \end{tabular}
    \caption{Hyperparameter setup for RoBERTa-base on GLUE}
    \label{table: glue param}
\end{table*}

\end{document}